\documentclass[11pt]{article}

\usepackage[utf8]{inputenc}
\usepackage[margin=1in]{geometry}
\usepackage{amsmath,amssymb}
\usepackage{graphicx}
\usepackage{booktabs}
\usepackage{multirow}
\usepackage{xcolor}
\usepackage{hyperref}
\hypersetup{hidelinks}
\usepackage{cleveref}
\usepackage{natbib}
\usepackage{subcaption}
\usepackage{tabularx}

\newcommand{\modelA}{RealSafe-R1-7B}
\newcommand{\modelB}{DeepSeek-R1-Distill-Qwen-7B}
\newcommand{\modelC}{saferlhf\_ultra\_sft}
\newcommand{\modelD}{Llama-3.1-8B}

\title{Localizing Safety Alignment: MLP Layers and Mid-Network Blocks Encode Refusal Behavior in Large Language Models}

\author{
  Mingyu Zong \thanks{The authors thank collaborators Jwala Dhamala and Rahul Gupta from Amazon for helpful directions and inputs.}\\
  Department of Computer Science\\
  University of Southern California \\
  Los Angeles, CA 90007 \\
  \texttt{mzong@usc.edu} \\
  \and
  Sampad Mohanty \\
  Department of Computer Science\\
  University of Southern California \\
  Los Angeles, CA 90007 \\
  \texttt{sbmohant@usc.edu} \\
  \and
  Bhaskar Krishnamachari \\
  Department of Electrical and Computer Engineering \\ 
  Department of Computer Science\\
  University of Southern California \\
  Los Angeles, CA 90007 \\
  \texttt{bkrishna@usc.edu} \\
}

\date{}

\begin{document}

\maketitle

\begin{abstract}
Safety alignment in large language models is often treated as a distributed property of the entire network, yet its practical brittleness suggests that refusal behavior may be concentrated in a smaller set of parameters. This work addresses where safety-aligned refusal is encoded by transplanting weights from aligned models into matched unaligned base models at multiple levels of granularity. Using two open-weight model pairs and four safety benchmarks, we conducted experiments to compare the effects of replacing attention weights, MLP weights, contiguous layer regions, and MLP blocks. Across both model families, refusal transfer is dominated by MLP weights: replacing MLP parameters recovers substantially more malicious-prompt refusal than replacing attention parameters, with gains of at least 2.7 times more across benchmarks. Within the MLP stack, refusal-relevant parameters exhibit a consistent mid-network concentration, as the block spanning layers 8–11 is selected first in all six greedy searches over model-dataset pairs. The results also show that the composition of safety-relevant components is non-additive: in five of six greedy trajectories, adding more aligned blocks can reduce refusal performance, and selective block subsets can outperform full MLP transplantation on malicious refusal, benign over-refusal, or both. Finally, greedy orders transferred to OR-Bench vary with the source benchmark used to derive them, indicating a benchmark-dependent precision-coverage trade-off. These results suggest that safety alignment in current LLMs is both localized and interaction-sensitive, offering insight into alignment brittleness and potential avenues for targeted safety interventions. \footnote{Code and datasets are available at \url{https://github.com/ANRGUSC/Localizing-Safety-Alignment}.}
\end{abstract}

\section{Introduction}
Large language models (LLMs) are Transformer-based neural language models trained on large-scale text corpora to predict and generate natural language outputs \citep{vaswani2017attention, brown2020languagemodelsfewshotlearners}. Scaling model size, dataset size, and computational resources has led to broad improvements in language modeling and downstream task performance, making LLMs increasingly capable general-purpose agents \citep{kaplan2020scalinglawsneurallanguage, bommasani2022opportunitiesrisksfoundationmodels}. In practice, many deployed LLMs are first pretrained to acquire general linguistic and world knowledge, then adapted to downstream tasks so that their responses better match user intent, task requirements, and social norms. This two-stage development process makes safety alignment especially important: pre-training provides broad capabilities, but whether and when a model refuses harmful requests is largely shaped during post-training.

Modern alignment pipelines, including supervised instruction tuning, reinforcement learning from human feedback (RLHF), and preference-based optimization, have substantially improved safety of LLMs \citep{amballa2024safeservealigninginstructiontuned, ouyang2022training, bai2022traininghelpfulharmlessassistant, rafailov2023direct}. However, aligned refusal behavior tends to be brittle in practice. Prior work has shown that small downstream updates can weaken safeguards or unintentionally alter safety behavior \citep{Qi2023FinetuningAL, Yang2023ShadowAT}. In addition, carefully designed attacks such as jailbreaks, adversarial suffixes, automated red-teaming, and prompt transformations can bypass refusal mechanisms, suggesting that refusal is not a fixed property of model capability but a learned behavior sensitive to prompt format, model family, and internal representation \citep{Wei2023JailbrokenHD, Zou2023UniversalAT, perez2022redteaminglanguagemodels, carlini2024alignedneuralnetworksadversarially, Zhang2026ExposingLS}. At the same time, safety tuning can also produce the opposite failure mode, which is over-refusal on harmless or legitimate prompts \citep{Bianchi2023SafetyTunedLL, rottger2024xstest}. Therefore, LLM safety must be evaluated along two axes: resistance to harmful instructions and avoidance of over-refusal on harmless prompts. A model that refuses too rarely remains vulnerable to misuse, while a model that refuses too broadly may become unhelpful.

These observations motivate a more structured understanding of how refusal is implemented inside LLMs. A growing body of mechanistic work demonstrates that safety-relevant behaviors can often be detected or steered through internal activations. Activation space studies have identified low-dimensional directions associated with refusal behavior \citep{Arditi2024RefusalIL, Wollschlager2025TheGO}. Another line of work utilizes localized parameter or representation interventions to show that targeted manipulations can produce predictable behavioral changes \citep{Wei2024AssessingTB, Zhao2024DefendingLL, Lee2024ProgrammingRW}. These findings indicate that refusal is not uniformly distributed throughout the network, but they leave open a complementary weight-space question: which parameters, and at which depths, are responsible for transferring safety refusal from an aligned model to its base counterpart?

We hypothesize that safety-aligned refusal is localized: the parameters mediating transferable refusal behavior are concentrated in specific weight matrices and depth ranges rather than uniformly distributed across the network. Such concentration could help explain the brittleness of aligned refusal under targeted parameter updates and could enable more selective interventions. We test this hypothesis through selective weight transplantation. Using two matched open-weight model pairs: \modelA{}/\modelB{} and \modelC{}/\modelD{}, we construct hybrid models by replacing subsets of parameters in the unaligned base model with the corresponding parameters from the aligned model. Because each pair shares the same architecture, behavioral differences can be attributed directly to the transplanted weights rather than to model design. By varying the granularity of transplantation, we answer these finer localization questions: Is transferable refusal behavior mediated more strongly by attention weights or MLP weights? Is it concentrated in particular depth ranges? 

Our experiments provide consistent evidence across both model families that transferable refusal behavior is mediated more strongly by MLP than attention weights. Within the MLP pathway, a mid-depth block spanning layers 8–11 is selected first in all six greedy searches over model-dataset pairs, indicating a consistent concentration of refusal-relevant parameters at this depth. At the same time, aligned blocks do not compose additively: in most settings, adding additional aligned blocks does not guarantee increased refusal, and selective MLP-block subsets can outperform full MLP transplantation. We further validate these patterns using larger filtered subsets. To sum up, this paper suggests that safety-aligned refusal behavior in the model families studied is largely mediated by MLP weights and is particularly concentrated in mid-network layers, based on empirical evidence.

The rest of the paper is organized as follows: the next section introduces related work on safety alignment. \Cref{sec:experimental-setup} describes the matched model pairs, hybrid models' construction procedures, evaluation benchmarks, and greedy block selection method used to localize safety-relevant parameters, followed by the results section which presents the main findings. \Cref{sec:discussion} discusses the implications of these findings for alignment brittleness, safety precision, and benchmark choice. Finally, we conclude with limitations of this study and directions for future work.

\section{Related Work}

Safety alignment in large language models is commonly achieved through supervised instruction tuning, reinforcement learning from human feedback, constitutional AI, or preference-based optimization methods. Foundational work has established instruction tuning and RLHF as practical mechanisms for training models to follow user instructions while incorporating human preferences over helpfulness and harmlessness \citep{ouyang2022training, bai2022traininghelpfulharmlessassistant}. Constitutional AI further reduces reliance on direct human labels by using rule-based self-critique and AI feedback to improve harmlessness \citep{Bai2022ConstitutionalAH}. More recent preference-optimization methods such as Direct Preference Optimization (DPO) simplify the RLHF pipeline by optimizing directly on preference pairs without explicitly training a separate reward model \citep{rafailov2023direct}. Building on this paradigm, safety-focused instruction-tuning methods directly optimize refusal behavior on harmful requests in open-weight models \citep{amballa2024safeservealigninginstructiontuned}. These approaches demonstrate that refusal behavior can be reliably induced through post-training, but they also imply that refusal is learned and may therefore be sensitive to later updates or changes in prompting. 

Despite the effectiveness of post-training alignment, evaluations on aligned LLMs show that learned refusal is not fully robust. Downstream adaptation affects safety, showing that fine-tuning or other targeted updates can substantially weaken aligned safeguards \citep{Qi2023FinetuningAL, Yang2023ShadowAT}. Automated red-teaming methods can instruct language models themselves to discover prompts that elicit undesirable outputs \citep{perez2022redteaminglanguagemodels}. Other attacks search for adversarial suffixes or transform harmful instructions into alternative formats, including mathematical or encoded prompts, in order to bypass safety filters \citep{Zou2023UniversalAT, carlini2024alignedneuralnetworksadversarially, Zhang2026ExposingLS}. These attacks highlight that many refusal mechanisms are sensitive to surface form and distribution shift. On the other hand, excessive refusal is also problematic: models may reject benign prompts that contain sensitive terms or resemble unsafe requests, reducing their practical usefulness \citep{Bianchi2023SafetyTunedLL, rottger2024xstest, Cui2024ORBenchAO}.

In order to understand and address the brittleness of safety alignment, recent work studies how safety-related behavior is organized within the internal structure of language models. \citet{Arditi2024RefusalIL} claim that, for each model they study, a single extracted direction strongly mediates refusal. They also demonstrate that ablating that direction suppresses refusal, while adding it to the model induces over-refusal on benign inputs. \citet{Wollschlager2025TheGO} argue that refusal does not reside in a single direction, but is organized in multi-dimensional polyhedral cones that contain multiple refusal-mediating directions. Despite their disagreement about geometry, both papers suggest that refusal is not uniformly encoded across the network, but depends on structured and manipulable internal features. Furthermore, prior work also proves that targeted interventions can predictably alter a model's safety behavior. Using layer-wise pruning, \citet{Zhao2024DefendingLL} identify ``safety layers'' inside a model. The authors then propose Layer-specific Editing (LED) to restrict updates on those layers so that downstream toxic layers decode toward safe refusals instead of harmful continuations. \citet{Lee2024ProgrammingRW} propose Conditional Activation Steering (CAST), which applies a refusal steering vector when a prompt's hidden states match a learned condition vector, thus enabling selective refusal rules such as refusing only particular categories of content while preserving normal responses for others. This result further supports the view that refusal can be controlled through compact internal representations. At the parameter level, \citet{Wei2024AssessingTB} study safety brittleness through pruning and low-rank modifications. Critical neurons and ranks are distinguished by disentangling safety-related components. Surprisingly, the safety-critical regions are sparse, taking up around 3\% of parameters and 2.5\% of ranks. It is also reported that safety and utility appear more differentiated in MLP layers than in attention layers, which is in line with the findings from this work.

Beyond internal evaluation, a large number of benchmarks has been developed to measure both resistance to harmful requests and failure on benign ones. AdvBench tests whether models comply with harmful instructions under adversarial prompting \citep{Zou2023UniversalAT}, while TwinPrompt and SGXSTest extend this style of evaluation with paired settings designed to probe safety behavior more systematically \citep{Krau2025TwinBreakJL, Gupta2024WalledEvalAC}. Complementary benchmarks emphasize the importance of false positives. XSTest evaluates exaggerated safety behavior on safe prompts that resemble unsafe ones \citep{rottger2024xstest}. OR-Bench evaluates whether aligned models over-refuse harmless prompts \citep{Cui2024ORBenchAO}.

Existing work has primarily characterized safety at two levels: as an internal phenomenon studied through activation-level analysis or targeted editing, and as an external behavioral property measured by benchmarks. Less attention has been drawn to a direct weight-space localization question: in a safety-aligned model, which parameters actually carry the transferable refusal behavior relative to its unaligned counterpart? This paper addresses this question by comparing selective transplants of attention weights, MLP weights, contiguous layer ranges, and MLP blocks across matched base and aligned model pairs. The experiments are set up to connect safety evaluation with mechanistic localization, focusing specifically on where aligned refusal behavior resides in the network.

\section{Experimental Setup}
\label{sec:experimental-setup}

\begin{table}[htbp]
\centering
\begin{tabular}{|p{5.5cm}|c|c|p{1.7cm}|p{1.7cm}|}
\hline
\multicolumn{2}{|c|}{\textbf{Models}} & \multicolumn{3}{c|}{\textbf{Benchmarks}} \\
\hline
\textbf{Aligned/Base Pair} & \textbf{Layers} & \textbf{Name} & \textbf{Harmful Prompts} & \textbf{Benign Prompts} \\
\hline
\multirow{2}{=}{RealSafe-R1-7B / DeepSeek-R1-Distill-Qwen-7B
} & \multirow{2}{*}{28} & TwinPrompt & \checkmark & \checkmark \\
                         &                      & SGXSTest & \checkmark & \checkmark \\
\cline{1-2}
\multirow{2}{=}{saferlhf\_ultra\_sft / Llama-3.1-8B} & \multirow{2}{*}{32} & AdvBench & \checkmark &  \\
                         &                      & OR-Bench &  & \checkmark \\
\hline
\end{tabular}
\caption{Overview of models and benchmarks.}
\label{tab: model_benchmark}
\end{table}

\subsection{Model Pairs}
We analyze two pairs of open-weight models that share the same architecture within each pair but differ in whether they were safety-aligned (Table \ref{tab: model_benchmark}). The first pair consists of \modelA{} and \modelB{}, both are 28-layer Qwen2-based models \citep{Suma2025DeepSeekR1IR, Zhang2025RealSafeR1SD}. \modelA{} underwent safety training on approximately 15K examples, including roughly 10K direct harmful queries and 5K jailbreak prompts. The second pair consists of \modelC{} and \modelD{} \citep{huang2025saferlhf_ultra_sft, grattafiori2024llama}. Both models are 32-layer Llama models, with \modelC{} being the fine-tuned version on the instruct variant of \modelD{} on the saferlhf\_ultra dataset \citep{ji2025pku}.

\subsection{Weight Transplantation Procedure}
All hybrid models are constructed by replacing selected weight matrices in the base model with the corresponding matrices from the aligned model. Because of the shared architecture, behavioral differences in the resulting hybrids can be attributed directly to the transplanted weights.

\subsection{Granularity of Interventions}
This study evaluates several levels of transplantation granularity.

First, we study component-level hybrids. The \texttt{attn} configuration replaces all attention projections (\texttt{q\_proj}, \texttt{k\_proj}, \texttt{v\_proj}, and \texttt{o\_proj}) across all layers, while keeping the base model's MLP weights unchanged. The \texttt{mlp} configuration performs the complementary intervention, replacing only the feed-forward projections (\texttt{gate\_proj}, \texttt{up\_proj}, and \texttt{down\_proj}). 

We also evaluate contiguous layer groups. The \texttt{first5}, \texttt{mid5}, and \texttt{last5} configurations transplant both attention and MLP weights for five consecutive layers, while \texttt{first\_half} and \texttt{second\_half} transplant the corresponding half of the network weights.

Motivated by the strong performance of \texttt{mlp} hybrids, we then localize safety within the MLP stack. We partition the model's MLP weights into contiguous blocks, each containing 4 layers. For \modelA{} with 28 layers, this yields seven blocks: B1 = layers 0-3 through B7 = layers 24-27. For \modelC{} with 32 layers, this yields eight blocks: B1 = layers 0-3 through B8 = layers 28-31. We first perform transplantation for each individual block and then analyze combinations of blocks.

\subsection{Evaluation Benchmarks and Subset Construction}
We conduct model evaluations on four benchmarks that probe complementary aspects of safety behavior: TwinPrompt \citep{Krau2025TwinBreakJL}, SGXSTest \citep{Gupta2024WalledEvalAC}, AdvBench \citep{Zou2023UniversalAT}, and OR-Bench \citep{Cui2024ORBenchAO}. TwinPrompt and SGXSTest provide paired harmful and benign prompts, allowing us to evaluate both refusal of unsafe requests and over-refusal of safe requests. AdvBench contains harmful instructions only and is used to measure malicious-prompt-refusal. OR-Bench contains benign prompts designed specifically to stress-test over-refusal (Table \ref{tab: model_benchmark}).

For each malicious benchmark, we construct evaluation subsets by retaining prompts on which the safety-aligned model refuses while the corresponding base model complies. We keep up to 30 prompts per condition. For the benign subsets used to measure over-refusal, we apply the analogous filtering procedure to the paired benchmarks. Because this filtering is more restrictive for some settings, certain benign subsets contain fewer than 30 examples; we clarify the total number of prompts when reporting results.

In addition, we construct a separate paired prompt pool and apply the same aligned–base disagreement filtering criterion to obtain larger validation subsets that contain 100 malicious and 100 benign prompts after filtering. So they preserve the behavioral contrast used in the primary experiments while testing whether the observed localization patterns persist at a larger sample size.

\subsection{Metrics}
We define a \emph{refusal behavior} to be an explicit rejection induced by recognized harmful or unethical implications of the request. Refusal on malicious datasets (MR) is higher if a model incorporates better defense mechanisms against malicious use. For benign prompts, we report benign over-refusal (BOR) results, which are the numbers of benign prompts incorrectly turned down; lower BOR is preferred for all models. All model answers are manually inspected. We include generation configurations in section \ref{sec:appendix}.

\subsection{Greedy Block Selection}
\label{sec:greedy}
To study how refusal-relevant blocks interact, we perform greedy forward selection separately for each model-dataset pair on the malicious benchmarks. Starting from the base model, we iteratively add the MLP block from the remaining set that yields the largest gain in refusal until all blocks are transplanted. Besides refusals, a model can output other acceptable and safe answers, which provide no useful harmful information, even if it is not phrased as an explicit refusal. When multiple candidates tie on refusals, we select the one with more acceptable safe answers. This procedure produces both (i) a dataset-specific importance ordering over blocks and (ii) the best-performing block subset at each budget $k$. The orders collected from malicious benchmarks are transferred to OR-Bench for generalization testing. In addition, with Block~3 constantly being selected first, we further examine the importance of individual Block~3 layers and their combinations.

\section{Results}

We find three consistent patterns across both model families. First, MLP weights play a more essential role than attention weights in recovering refusal behavior. In addition, refusal-relevant parameters are not uniformly distributed across the MLP stack: Block~3 (layers 8--11) is the most consistently effective individual block for transferring refusal behavior, within which layer 8 stands out. Moreover, greedy block selection reveals a six-block substitution that often outperforms full MLP transplantation on malicious refusal, benign over-refusal, or both.

\subsection{Component-Level Hybrids}
\label{sec:mlp-dominance}

\begin{table}[t]
\centering
\begin{tabular}{@{}llccc|p{1.5cm}@{}}
\toprule
\textbf{Model} & \textbf{Config} & \textbf{TwinPrompt} & \textbf{SGXSTest} & \textbf{AdvBench} & \textbf{TwinPrompt BOR} \\
\midrule
\multirow{7}{*}{\modelA{}}
& first5 & 1/30 (3.33\%) & 3/30 (10\%) & 2/30 (6.67\%) & 0/30 (0\%) \\
& mid5 & 8/30 (26.67\%) & 2/30 (6.67\%) & 11/30 (36.67\%) & 0/30 (0\%) \\
& last5 & 4/30 (13.33\%) & 2/30 (6.67\%) & 7/30 (23.33\%) & 0/30 (0\%) \\
& first\_half & 8/30 (26.67\%) & 11/30 (36.67\%) & 10/30 (33.33\%) & 1/30 (3.33\%)  \\
& second\_half & 9/30 (30.00\%) & 11/30 (36.67\%) & 18/30 (60.00\%) & 1/30 (3.33\%)  \\
& attn & 5/30 (16.67\%) & 2/30 (6.67\%) & 10/30 (33.33\%) & 0/30 (0\%) \\
& mlp & \textbf{19/30 (63.33\%)} & \textbf{17/30 (56.67\%)} & \textbf{27/30 (90.00\%)} & 11/30 (36.67\%) \\
\midrule
\multirow{7}{*}{\modelC{}}
& first5 & 1/30 (3.33\%)  & 0/30 (0\%) & 0/30 (0\%) & 0/5 (0\%) \\
& mid5 & 0/30 (0\%) & 0/30 (0\%) & 0/30 (0\%) & 0/5 (0\%) \\
& last5 & 0/30 (0\%) & 0/30 (0\%) & 0/30 (0\%) & 0/5 (0\%) \\
& first\_half & 18/30 (60.00\%) & 2/30 (6.67\%) & 10/30 (33.33\%) & 0/5 (0\%) \\
& second\_half & 0/30 (0\%) & 2/30 (6.67\%) & 0/30 (0\%) & 0/5 (0\%) \\
& attn & 5/30 (16.67\%) & 0/30 (0\%) & 0/30 (0\%) & 0/5 (0\%) \\
& mlp & \textbf{21/30 (70.00\%)} & \textbf{4/30 (13.33\%)} & \textbf{17/30 (56.67\%)} & 0/5 (0\%) \\
\bottomrule
\end{tabular}
\caption{Component-level hybrids' performance, including MR on three malicious subsets and BOR on benign TwinPrompt prompts.}
\label{tab:component-comparison}
\end{table}

Our first set of analyses compares hybrid models that transplant broad classes of parameters across three malicious-prompt benchmarks to lay the groundwork for finer localizations. We summarize MR and BOR from multiple benchmarks in Table \ref{tab:component-comparison}. The pattern is consistent across both model pairs: replacing MLP weights recovers substantially more safety behavior than replacing attention weights.

Under MLP weight transplantation, hybrid models from the \modelA{} family achieve 19, 17, and 27 refusals on 30-malicious-prompt subsets from TwinPrompt, SGXSTest, and AdvBench, respectively, which are the highest MR values among all component-level hybrids. Models from the \modelC{} family also yield best performance with this configuration. When compared with MR results generated by \texttt{attn} hybrids, MLP-only transplantation exceeds by at least 2.7 times across the six comparisons.

On the other hand, the same intervention that restores refusal behavior can also import over-refusal. For \modelA{} hybrids, full MLP substitution produces 11 benign refusals on TwinPrompt, whereas attention-only substitution produces none. Proper refusal and false-positive refusal are therefore not cleanly separable at the component level in this model family: both are concentrated primarily in the MLP pathway. By contrast, \modelC{} exhibits no TwinPrompt over-refusal behavior under full MLP transplantation, suggesting that the aligned signal transferred from \modelC{} may be more selective.

Taken together, these results identify MLP weights as the dominant parameter pathway for transferring aligned refusal behavior in both model families. Since MLP substitution clearly dominates, we then ask whether safety is distributed uniformly across the MLP stack or concentrated in particular blocks.

\subsection{Individual MLP Blocks}
\label{sec:layer-localization}

\begin{table}[t]
\centering
\begin{tabular}{@{}lccccccc@{}}
\toprule
\textbf{Dataset} & \textbf{B1} & \textbf{B2} & \textbf{B3} & \textbf{B4} & \textbf{B5} & \textbf{B6} & \textbf{B7} \\
& \textbf{L0-3} & \textbf{L4-7} & \textbf{L8-11} & \textbf{L12-15} & \textbf{L16-19} & \textbf{L20-23} & \textbf{L24-27} \\
\midrule
TwinPrompt & 2 & 3 & \textbf{7} & 6 & 3 & 3 & 1 \\
SGXSTest & \textbf{3} & \textbf{3} & \textbf{3} & 0 & \textbf{3} & 2 & \textbf{3} \\
AdvBench & 4 & 7 & \textbf{10} & 7 & 9 & 5 & 5 \\
\bottomrule
\end{tabular}
\caption{Individual MLP-block substitution results for the \modelA{} family. Values represent refusal counts on malicious subsets (30 prompts per subset).}
\label{tab:individual-blocks-realsafe}
\end{table}

\begin{table}[t]
\centering
\begin{tabular}{@{}lcccccccc@{}}
\toprule
\textbf{Dataset} & \textbf{B1} & \textbf{B2} & \textbf{B3} & \textbf{B4} & \textbf{B5} & \textbf{B6} & \textbf{B7} & \textbf{B8} \\
& \textbf{L0-3} & \textbf{L4-7} & \textbf{L8-11} & \textbf{L12-15} & \textbf{L16-19} & \textbf{L20-23} & \textbf{L24-27} & \textbf{L28-31} \\
\midrule
TwinPrompt & 0 & 1 & \textbf{2} & 0 & 0 & 0 & 0 & 0 \\
SGXSTest & 0 & 0 & 0 & 0 & 0 & 0 & 0 & 0 \\
AdvBench & 0 & 0 & \textbf{1} & 0 & 0 & 0 & 0 & 0 \\
\bottomrule
\end{tabular}
\caption{Individual MLP-block substitution results for the \modelC{} family. Values represent refusal counts on malicious subsets (30 prompts per subset).}
\label{tab:individual-blocks-saferlhf}
\end{table}

In this follow-up experiment, we localize the safety signals within the MLP stack by transplanting one block at a time. Tables~\ref{tab:individual-blocks-realsafe} and~\ref{tab:individual-blocks-saferlhf} report MR results for each single-block substitution. Refusal-relevant effects are distributed across multiple MLP blocks, but Block~3 is the most consistently important block across benchmarks. On TwinPrompt and AdvBench, Block~3 achieves the highest MR rates for both model families, and on SGXSTest it appears among the top candidates. The recurrence of the same absolute depth range across the Qwen2-7B and Llama-3.1-8B architectures suggests that, for models of this scale, the parameters mediating transferable refusal behavior appear disproportionately concentrated around layers 8–11.

The results also reveal that individual MLP blocks in the \modelC{} family are generally less effective in rejecting unsafe instructions. A potential explanation is that both extent and variety of safety training affect the robustness of knowledge encoding, but verification is left for future research. Tables~\ref{tab:individual-blocks-realsafe} and~\ref{tab:individual-blocks-saferlhf} measure the effect of each block in isolation; \Cref{sec:composition} instead asks which blocks are most useful when added cumulatively.

\begin{table}[t]
\centering
\begin{tabular}{@{}lll@{}}
\toprule
\textbf{Model} & \textbf{Dataset} & \textbf{Greedy Order} \\
\midrule
\multirow{3}{*}{\modelA{}}
& TwinPrompt & \textbf{3}, 5, 2, 6, 4, 7, 1 \\
& SGXSTest & \textbf{3}, 6, 7, 4, 1, 5, 2 \\
& AdvBench & \textbf{3}, 6, 5, 1, 2, 7, 4 \\
\midrule
\multirow{3}{*}{\modelC{}}
& TwinPrompt & \textbf{3}, 4, 5, 2, 1, 8, 7, 6 \\
& SGXSTest & \textbf{3}, 7, 2, 4, 5, 1, 6, 8 \\
& AdvBench & \textbf{3}, 2, 4, 5, 7, 1, 8, 6 \\
\bottomrule
\end{tabular}
\caption{Greedy block orders on three malicious subsets. Block~3 is selected first in all six settings.}
\label{tab:greedy-orders}
\end{table}

\begin{figure}[htp]
\begin{subfigure}{\textwidth}
\centering
\includegraphics[width=0.85\textwidth,height=5.5cm]{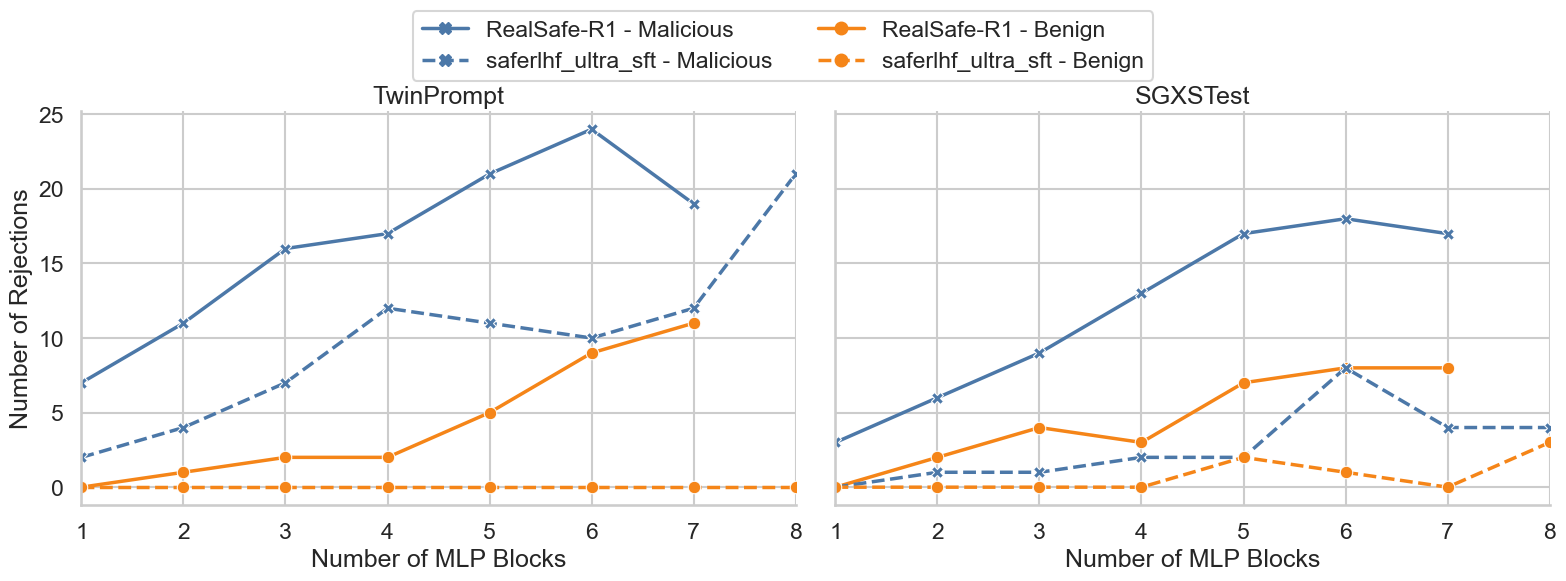}
\end{subfigure}
\bigskip
\begin{subfigure}{\textwidth}
\centering
\includegraphics[width=0.425\textwidth,height=4cm]{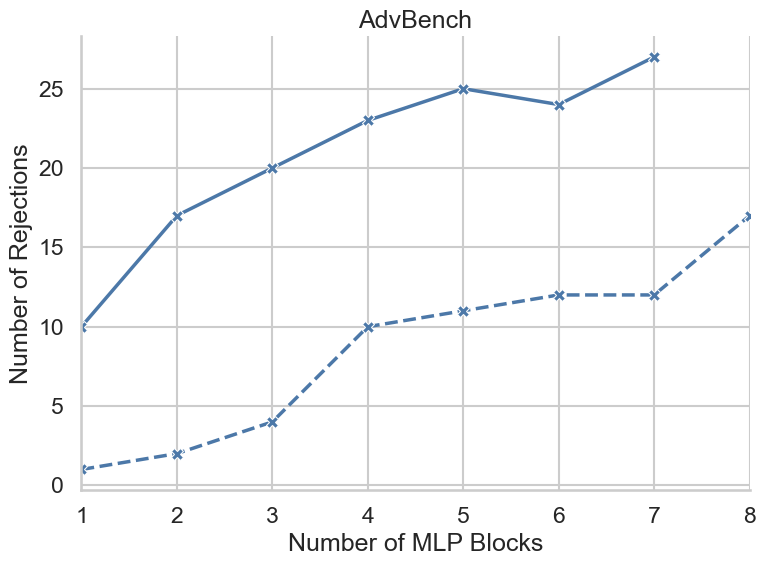}
\end{subfigure}
\caption{Hybrid model performance on benchmark subsets.}
\label{fig:hard_refusal_mal}
\end{figure}

\subsection{Non-monotonic Block Composition}
\label{sec:composition}

\begin{table}[t]
\centering
\begin{tabular}{@{}lp{1.5cm}p{1.5cm}p{1.5cm}p{1.5cm}p{1.5cm}p{1.5cm}p{1.5cm}p{1.5cm}@{}}
\toprule
\textbf{Dataset} & \textbf{k=1} & \textbf{k=2} & \textbf{k=3} & \textbf{k=4} & \textbf{k=5} & \textbf{k=6} & \textbf{k=7} \\
\midrule
\multicolumn{8}{l}{TwinPrompt} \\
\quad MR & 7/30 (23.33\%) & 11/30 (36.67\%) & 16/30 (53.33\%) & 17/30 (56.67\%) & 21/30 (70.00\%) & \textbf{24/30 (80.00\%)} & 19/30 (63.33\%) \\
\quad BOR & 0/30 (0\%) & 1/30 (3.33\%) & 2/30 (6.67\%) & 2/30 (6.67\%) & 5/30 (16.67\%) & 9/30 (30.00\%) & 11/30 (36.67\%) \\
\midrule
\multicolumn{8}{l}{SGXSTest} \\
\quad MR & 3/30 (10.00\%) & 6/30 (20.00\%) & 9/30 (30.00\%) & 13/30 (43.33\%) & 17/30 (56.67\%) & \textbf{18/30 (60.00\%)} & 17/30 (56.67\%) \\
\quad BOR & 0/30 (0\%) & 2/30 (6.67\%) & 4/30 (13.33\%) & 3/30 (10.00\%) & 7/30 (23.33\%) & 8/30 (26.67\%) & 8/30 (26.67\%) \\
\midrule
\multicolumn{8}{l}{AdvBench} \\
\quad MR & 10/30 (33.33\%) & 17/30 (56.67\%) & 20/30 (66.67\%) & 23/30 (76.67\%) & 25/30 (83.33\%) & 24/30 (80.00\%) & \textbf{27/30 (90.00\%)} \\
\bottomrule
\end{tabular}
\caption{Greedy trajectories for \modelA{} family. $k$ is the number of MLP blocks transplanted. Greedy orders: TwinPrompt 3-5-2-6-4-7-1; SGXSTest 3-6-7-4-1-5-2; AdvBench 3-6-5-1-2-7-4.}
\label{tab:greedy-realsafe}
\end{table}

\begin{table}[t]
\centering
\begin{tabular}{@{}lp{1.5cm}p{1.5cm}p{1.5cm}p{1.5cm}p{1.5cm}p{1.5cm}p{1.5cm}p{1.5cm}@{}}
\toprule
\textbf{Dataset} & \textbf{k=1} & \textbf{k=2} & \textbf{k=3} & \textbf{k=4} & \textbf{k=5} & \textbf{k=6} & \textbf{k=7} & \textbf{k=8} \\
\midrule
\multicolumn{9}{l}{TwinPrompt} \\
\quad MR & 2/30 (6.67\%) & 4/30 (13.33\%) & 7/30 (23.33\%) & 12/30 (40.00\%) & 11/30 (36.67\%) & 10/30 (33.33\%) & 12/30 (40.00\%) & \textbf{21/30 (70.00\%)} \\
\quad BOR & 0/5 (0\%) & 0/5 (0\%) & 0/5 (0\%) & 0/5 (0\%) & 0/5 (0\%) & 0/5 (0\%) & 0/5 (0\%) & 0/5 (0\%) \\
\midrule
\multicolumn{9}{l}{SGXSTest} \\
\quad MR & 0/30 (0\%) & 1/30 (3.33\%) & 1/30 (3.33\%) & 2/30 (6.67\%) & 2/30 (6.67\%) & \textbf{8/30 (26.67\%)} & 4/30 (13.33\%) & 4/30 (13.33\%) \\
\quad BOR & 0/18 (0\%) & 0/18 (0\%) & 0/18 (0\%) & 0/18 (0\%) & 2/18 (11.11\%) & 1/18 (5.56\%) & 0/18 (0\%) & 3/18 (16.67\%) \\
\midrule
\multicolumn{9}{l}{AdvBench} \\
\quad MR & 1/30 (3.33\%) & 2/30 (6.67\%) & 4/30 (13.33\%) & 10/30 (33.33\%) & 11/30 (36.67\%) & 12/30 (40.00\%) & 12/30 (40.00\%) & \textbf{17/30 (56.67\%)} \\
\bottomrule
\end{tabular}
\caption{Greedy trajectories for \modelC{} family. $k$ is the number of MLP blocks transplanted. Greedy orders: TwinPrompt 3-4-5-2-1-8-7-6; SGXSTest 3-7-2-4-5-1-6-8; AdvBench 3-2-4-5-7-1-8-6.}
\label{tab:greedy-saferlhf}
\end{table}

We run greedy forward selection to determine how safety-aligned blocks interact when added cumulatively. Block~3 is selected first in all six model-dataset pairs (Table~\ref{tab:greedy-orders}), consistent with its strong individual performance from previous experiment. Tables~\ref{tab:greedy-realsafe} and~\ref{tab:greedy-saferlhf} report MR and BOR results at each block budget~$k$. The main finding is that on two of three benchmarks, a six-block transplantation can outperform the full MLP transplantation on rejecting malicious requests.

The greedy trajectories show how this pattern develops across block budgets (Figure~\ref{fig:hard_refusal_mal}). For \modelA{} on TwinPrompt, MR rises from $7$ at $k=1$ to $24$ at $k=6$, then falls to $19$ at $k=7$. The trajectory of \modelC{} reaches $12$ MR counts at $k=4$, drops down to $10$ at $k=6$, and finally recovers when all eight blocks are present. On SGXSTest, \modelA{} peaks at MR $=18$ at $k=6$ and declines to $17$ at $k=7$. The drop observed from the \modelC{} family is more pronounced: performance peaks at $8$ at $k=6$ and falls to $4$ after the seventh block is added. AdvBench is the most monotonic setting. For \modelA{}, MR increases through $k=5$, dips at $k=6$, and recovers at $k=7$. The \modelC{}-AdvBench trajectory is the only fully monotonic one among the six, it starts with 1 refusal at $k=1$ and gradually grows to 17 rejections.

The BOR results further show benefits of using only six blocks of MLP weights. In the \modelA{}-TwinPrompt setting, the six-block hybrid achieves MR $=24$ with BOR $=9$, whereas the full seven-block transplant yields MR $=19$ with BOR $=11$, which is worse on both metrics. For \modelA{}-SGXSTest, full transplantation loses one refusal without reducing over-refusal. For \modelC{}-SGXSTest, the contrast is sharper: the six-block model achieves MR $=8$ with BOR $=1$, compared with MR $=4$ and BOR $=3$ for the full eight-block model.

We refer to the pattern where adding an aligned block reduces MR as the negative marginal effect. Five of the six trajectories exhibit such effect, but the disruptive block varies across settings: the first decline on both TwinPrompt trajectories coincides with the introduction of Block~1, whereas the sharp SGXSTest drop for \modelC{} is associated with Block~6. This dataset dependence stands against a single universally harmful block and instead supports that block utility is contingent on the surrounding set of already-transplanted blocks.

\begin{table}[t]
\centering
\begin{tabular}{@{}lccccccc@{}}
\toprule
\textbf{Model / Order} & \textbf{k=1} & \textbf{k=2} & \textbf{k=3} & \textbf{k=4} & \textbf{k=5} & \textbf{k=6} & \textbf{k=7} \\
\midrule
\multicolumn{8}{l}{\modelA{}} \\
\quad TwinPrompt order & \textbf{0} & \textbf{0} & 1 & \textbf{0} & 2 & 4 & \textbf{3} \\
\quad SGXSTest order & \textbf{0} & 1 & \textbf{0} & 1 & \textbf{0} & \textbf{3} & \textbf{3} \\
\quad AdvBench order & 1 & 2 & 3 & 3 & 1 & 4 & 6 \\
\midrule
\multicolumn{8}{c}{} \\
\end{tabular}

\begin{tabular}{@{}lcccccccc@{}}
\toprule
\textbf{Model / Order} & \textbf{k=1} & \textbf{k=2} & \textbf{k=3} & \textbf{k=4} & \textbf{k=5} & \textbf{k=6} & \textbf{k=7} & \textbf{k=8} \\
\midrule
\multicolumn{9}{l}{\modelC{}} \\
\quad TwinPrompt order & \textbf{0} & \textbf{0} & \textbf{0} & 1 & 1 & 1 & 1 & \textbf{3} \\
\quad SGXSTest order & \textbf{0} & \textbf{0} & \textbf{0} & \textbf{0} & \textbf{0} & \textbf{0} & \textbf{0} & \textbf{3} \\
\quad AdvBench order & \textbf{0} & 1 & 1 & 1 & 1 & 1 & 1 & \textbf{3} \\
\bottomrule
\end{tabular}
\caption{OR-Bench over-refusal (out of 30 benign prompts) under different greedy block orders. Lower is better. Bold indicates the best value at each budget within a model family.}
\label{tab:orbench-transfer}
\end{table}

\subsection{Transfer to OR-Bench is Benchmark-dependent}
\label{sec:generalization}

The greedy selection approach offers insights into MR-BOR trade-off with respect to model-dataset pairs. However, the BOR results are collected from datasets with paired prompts that minimize linguistic difference. We further evaluate the models and orders on a separate benign-prompt only benchmark, namely OR-Bench, to determine transferability and whether a specific order outperforms the others. As a result, the AdvBench-derived order is generally more conservative and often incurs higher OR-Bench over-refusal rates. By contrast, the order derived from SGXSTest tends to yield lower OR-Bench over-refusal than the order derived from AdvBench. For \modelA{}, the SGXSTest order attains the lowest or tied-lowest BOR at five of the seven block budgets. For \modelC{}, the SGXSTest order maintains zero over-refusals through most $k$ values. Nonetheless, this comparison should be interpreted cautiously. Regardless of the lower over-refusal values, SGXSTest-derived orders also tend to achieve lower malicious-prompt rejection than orders derived from TwinPrompt or AdvBench. We therefore interpret these results as evidence of a trade-off rather than a strict improvement. The source benchmark influences both how aggressively safety transfers and how much that transfer over-generalizes to benign prompts.

A second pattern we discover is that \modelC{} hybrids remain more precise than \modelA{} hybrids under transfer. At high budgets, \modelA{} incurs 3-6 BOR at $k=7$, depending on the order, whereas \modelC{} incurs 0-1 refusals at $k=7$ and reaches only 3 when all MLP blocks are transplanted. This suggests that the aligned signal inherited from \modelC{} is less prone to over-generalization under block composition.

Overall, the transfer results demonstrate that greedy block orders learned from different malicious benchmarks behave differently on unseen benign prompts, thus, benchmark choice matters for downstream precision.

\subsection{A Peek into Block~3}
\begin{table}[ht]
\centering
\setlength{\tabcolsep}{10pt}
\begin{tabular}{c|c|c|c|c|c|c}
\hline
\multicolumn{7}{c}{\textbf{TwinPrompt}} \\
\hline
\textbf{Layers} & \textbf{8} & \textbf{9} & \textbf{10} & \textbf{11} & & \\ 
\textbf{MR} & 3 & 1 & 2 & 1 & & \\
\hline
\textbf{Layers} & \textbf{8, 9} & \textbf{8, 10} & \textbf{8, 11} & \textbf{9, 10} & \textbf{9, 11} & \textbf{10, 11} \\ 
\textbf{MR} & 2 & 7 & 4 & 4 & 3 & 4 \\
\hline
\textbf{Layers} & \textbf{8, 9, 10} & \textbf{8, 9, 11} & \textbf{8, 10, 11} & \textbf{9, 10, 11} & & \\
\textbf{MR} & 4 & 3 & 2 & 6 & & \\
\hline
\textbf{Layers} & \textbf{Block3} & & & & & \\ 
\textbf{MR} & 7 & & & & & \\
\hline
\end{tabular}
\par\vspace{0.15cm}\par
\begin{tabular}{c|c|c|c|c|c|c}
\hline
\multicolumn{7}{c}{\textbf{SGXS}} \\
\hline
\textbf{Layers} & \textbf{8} & \textbf{9} & \textbf{10} & \textbf{11} & & \\ 
\textbf{MR} & 5 & 3 & 1 & 4 & & \\
\hline
\textbf{Layers} & \textbf{8, 9} & \textbf{8, 10} & \textbf{8, 11} & \textbf{9, 10} & \textbf{9, 11} & \textbf{10, 11} \\ 
\textbf{MR} & 3 & 0 & 4 & 2 & 2 & 3 \\
\hline
\textbf{Layers} & \textbf{8, 9, 10} & \textbf{8, 9, 11} & \textbf{8, 10, 11} & \textbf{9, 10, 11} & & \\
\textbf{MR} & 3 & 4 & 2 & 3 & & \\
\hline
\textbf{Layers} & \textbf{Block3} & & & & & \\ 
\textbf{MR} & 3 & & & & & \\
\hline
\end{tabular}
\par\vspace{0.15cm}\par
\begin{tabular}{c|c|c|c|c|c|c}
\hline
\multicolumn{7}{c}{\textbf{AdvBench}} \\
\hline
\textbf{Layers} & \textbf{8} & \textbf{9} & \textbf{10} & \textbf{11} & & \\ 
\textbf{MR} & 6 & 5 & 4 & 5 & & \\
\hline
\textbf{Layers} & \textbf{8, 9} & \textbf{8, 10} & \textbf{8, 11} & \textbf{9, 10} & \textbf{9, 11} & \textbf{10, 11} \\ 
\textbf{MR} & 2 & 3 & 6 & 4 & 4 & 4 \\
\hline
\textbf{Layers} & \textbf{8, 9, 10} & \textbf{8, 9, 11} & \textbf{8, 10, 11} & \textbf{9, 10, 11} & & \\
\textbf{MR} & 8 & 5 & 10 & 6 & & \\
\hline
\textbf{Layers} & \textbf{Block3} & & & & & \\ 
\textbf{MR} & 10 & & & & & \\
\hline
\end{tabular}
\caption{Refusal behaviors of RealSafe hybrid models constructed using Block~3 layers. All subsets contain 30 prompts.}
\label{tab:refusal_block3_layers}
\end{table}

Given the consistently strong performance of Block~3 weights, we build several hybrid models using only Block~3 layers to test if the non-monotonic pattern and marginal effect exist at a finer-grained level. As shown in Table~\ref{tab:refusal_block3_layers}, exploratory analysis further points to layer 8 for the \modelA{} family. Layer 8 produces more malicious-prompt refusals than the other individual layers. Transplanting only this aligned layer of weights to the base DeepSeek-R1-Distill-Qwen-7B model gives near or above half of the MR that the full block achieves. The exploratory evidence indicates that transferable refusal behavior is concentrated in this layer within Block~3. Across all three benchmarks, we observe the non-additivity from the combination of layers 8 and 9, along with a few others that are dataset-specific. Another consistency is that the full-block performance can be achieved or even surpassed by a subset of layers. We conduct the same experiments on the \modelC{} pair. Despite Block~3 being the most important MLP block, it induces at most 2 rejections on the subsets, so the results fail to reveal valuable information and are excluded from display.

\subsection{Validation on a Larger Paired Subsets}
\label{sec:validation}

\begin{table}[t]
\centering
\begin{tabular}{llcc}
\toprule
\textbf{Model} & \textbf{Config} & \textbf{MR} & \textbf{BOR} \\
\midrule
\multirow{7}{*}{\modelA{}}
& first5 & 8 & 0 \\
& mid5 & 8 & 0 \\
& last5 & 8 & 0 \\
& first\_half & 24 & 4 \\
& second\_half & 44 & 7 \\
& attn & 8 & 2 \\
& mlp & \textbf{61} & \textbf{33}\\
\bottomrule
\end{tabular}
\caption{Component-level hybrids' performance on validation subsets.}
\label{tab:component-level-results}
\end{table}

\begin{table}[t]
\centering
\begin{tabular}{@{}lccccccc@{}}
\toprule
\textbf{} & \textbf{B1} & \textbf{B2} & \textbf{B3} & \textbf{B4} & \textbf{B5} & \textbf{B6} & \textbf{B7} \\
& \textbf{L0-3} & \textbf{L4-7} & \textbf{L8-11} & \textbf{L12-15} & \textbf{L16-19} & \textbf{L20-23} & \textbf{L24-27} \\
\midrule
MR & 4 & 8 & \textbf{13} & 7 & 7 & 11 & 7 \\
BOR & 0 & 0 & \textbf{1} & 0 & 0 & 0 & \textbf{1} \\
\bottomrule
\end{tabular}
\caption{Individual MLP-block substitution results for the \modelA{} family.}
\label{tab:mlp-blocks-validation}
\end{table}

\begin{table}[t]
\centering
\begin{tabular}{@{}lp{1.5cm}p{1.5cm}p{1.5cm}p{1.5cm}p{1.5cm}p{1.5cm}p{1.5cm}p{1.5cm}@{}}
\toprule
\textbf{} & \textbf{k=1} & \textbf{k=2} & \textbf{k=3} & \textbf{k=4} & \textbf{k=5} & \textbf{k=6} & \textbf{k=7} \\
\midrule
\quad MR & 13 & 29 & 42 & 52 & 54 & \textbf{64} & 61 \\
\quad BOR & 1 & 2 & 7 & 14 & 18 & 23 & \textbf{33} \\
\bottomrule
\end{tabular}
\caption{Greedy trajectories for \modelA{} family. $k$ is the number of MLP blocks transplanted. Greedy orders: Malicious Subset 3-5-6-7-2-4-1; Benign Subset 3-4-7-6-1-2-5.}
\label{tab:greedy-validation}
\end{table}

The preceding experiments rely on relatively small filtered subsets, with at most 30 prompts per condition. To test whether the observed localization patterns are robust to a larger sample of aligned–base disagreement cases, we conduct an additional validation experiment using a newly constructed paired prompt set. We created a list of paired actions, and randomly draw prefixes from another set to form complete instructions. We then followed the same subset construction criterion. For malicious prompts, we retained examples for which the safety-aligned \modelA{} model refuses while the corresponding \modelB{} base model complies. For benign prompts, we applied the analogous procedure to identify cases relevant to over-refusal. We constructed a substantially larger pool of candidate prompts, allowing both the malicious and benign subsets to contain 100 prompts after filtering. Therefore the experiment preserves the behavioral contrast central to our transplantation analysis while reducing the sensitivity of the reported results to the small sample sizes.

We repeat the transplantation experiments on the \modelA{} model family using the same transplantation procedures as in the main experiments. In particular, we examine whether the three principal observations obtained from the original subsets persist at the larger sample size: the dominance of MLP over attention weights in transferring refusal behavior, the concentration of safety-relevant weights in the mid-network MLP blocks, and the non-additive behavior of MLP-block composition.

The results in Table~\ref{tab:component-level-results} strongly reproduce the dominance of the MLP pathway. On the 100 malicious prompts, the \texttt{mlp} hybrid refuses 61 prompts, compared with only 8 for the \texttt{attn} hybrid. The remaining coarse-grained transplantations are substantially weaker: \texttt{first5}, \texttt{mid5}, and \texttt{last5} each produce 8 refusals, while \texttt{first\_half} and \texttt{second\_half} transplantation produce 24 and 44 refusals, respectively. These results are consistent with the original experiments and provide additional evidence that transferable refusal behavior in the \modelA{} family is concentrated primarily in MLP rather than attention weights. The larger benign subset also reproduces the precision trade-off associated with MLP transplantation. The \texttt{mlp} hybrid over-refuses 33 of 100 benign prompts, compared with at most 7 over-refusals from other settings. Thus, the same parameter pathway that transfers the greatest amount of malicious-prompt refusal also transfers substantially more over-refusal, strengthening the evidence that desirable refusal and false-positive refusal are not cleanly separable at the component level in this model family.

We further repeat the individual MLP-block experiment on the larger subsets. Block 3 (layers 8–11) still produces the highest malicious-prompt refusal rates among all seven individual blocks, rejecting 13 of 100 malicious prompts, followed by Block 6 with 11 rejections and Block 2 with 8 rejections. The remaining blocks produce between 4 and 7 rejections. On the other hand, Block 3 induces only 1 refusal among the 100 benign prompts (Table~\ref{tab:mlp-blocks-validation}). The recurrence of Block 3 as the strongest individual MLP block on a substantially larger prompt set is consistent with the mid-network localization observed in the original experiments.

Following greedy forward selection, malicious-prompt refusals increase from 13 with one block to 29, 42, 52, 54, and a peak of 64 as the block budget increases from $k=1$ to $k=6$. However, transplanting the final remaining MLP block reduces performance from 64 to 61 refusals at $k=7$ (Table~\ref{tab:greedy-validation}). Because the seven-block configuration is equivalent to full MLP transplantation, this result once more demonstrates that transplanting all aligned MLP weights need not maximize refusal behavior. The persistence of a negative marginal effect on a substantially larger prompt set supports the conclusion that refusal-relevant MLP blocks do not compose strictly additively.

Overall, the larger-subset validation reproduces all three principal localization patterns for the \modelA{} family: MLP weights dominate attention weights in transferring refusal behavior, Block 3 remains the strongest individual MLP block, and selective block composition can outperform full MLP transplantation. These results reduce the likelihood that the original observations are artifacts of the small evaluation subsets. However, because the validation examples are constructed using the same aligned–base disagreement criterion, this experiment establishes robustness to increased sample size within the filtered evaluation distribution rather than generalization to an unfiltered prompt distribution.

\section{Discussion}
\label{sec:discussion}

\subsection{Weight-Space Localization of Refusal Behavior}
Our transplantation experiments indicate that the parameters mediating safety-aligned refusal are not distributed uniformly across the network. Across both model families, transferable refusal behavior is mediated predominantly by the MLP pathway, with a mid-depth region spanning layers 8–11 repeatedly prioritized. Even within the mid-depth region, individual layers bear different levels of refusal-related behavior. This concentration provides a plausible explanation for the brittleness of alignment reported in prior work \citep{Qi2023FinetuningAL, Yang2023ShadowAT}. Since a relatively small subset of parameters carries a disproportionate share of the refusal signal, targeted updates to that subset can substantially alter safety behavior.

Moreover, our transplantation experiments localize \emph{where} safety-relevant information is stored, rather than \emph{how} it is computed. In particular, the critical role of MLP layers and the highlighted importance of Block~3 point to a narrow part of the network that merits deeper mechanistic analysis. We plan to combine the present weight-level interventions with activation-level methods, such as activation patching, causal tracing, or neuron-level probing, to characterize the representations and transformations within these blocks that support refusal behavior as a follow-up study.

\subsection{Interactions and Precision Trade-offs}
The greedy trajectories show that aligned blocks do not contribute independently. Adding an additional block can improve refusals in one context while degrading them in another, and the full MLP transplant is often not the best operating point. Thus, MLP-block interactions appear to be non-additive.

Another observation is that the same concentration that enables safety transfer can also introduce over-refusal. With respect to \modelA{} hybrids, for example, the MLP-only transplant simultaneously increases malicious prompt refusal and benign over-refusal. This co-localization means that improving safety precision is not only a matter of transplanting more aligned parameters; it requires identifying and choosing subsets whose interactions preserve the distinction between harmful and benign inputs.

More broadly, the findings suggest two concrete directions for alignment research. The first is \emph{precision-oriented alignment}: rather than optimizing refusal strength alone, future methods should explicitly target the precision-coverage trade-off that becomes visible in our transplantation experiments. The second is \emph{distributed robustness}: if current safety behavior depends heavily on a small number of MLP blocks, then a promising goal is to develop alignment procedures that encode safety more redundantly across the network, making it less brittle to fine-tuning, editing, or weight replacement.

\subsection{Benchmark Choice and Deployment Implications}
Our OR-Bench transfer results indicate that the benchmark used to derive the greedy order materially affects downstream behavior on unseen benign prompts. Some source benchmarks induce more aggressive safety transfer, which can improve malicious-prompt rejection but also increase over-refusal on benign prompts. Others induce a milder transfer that preserves benign compliance more effectively, although at the cost of lower malicious refusal. In this sense, the observed differences are better understood as a precision-coverage trade-off than as evidence that one benchmark is categorically superior. 

This trade-off is especially important for deployment. If the goal is to maximize refusal on harmful prompts, a more aggressive order may be preferable even if it incurs higher BOR. On the other hand, if the goal is to preserve usability on benign inputs, a milder order may be preferable, even if it leaves more harmful prompts answered unsafely. In practice, block selection should not be evaluated on malicious-prompt rejection alone: the choice of source benchmark can materially affect both safety strength and over-generalization.

\subsection{Limitations and Future Directions}
This study provides evidence that safety alignment is localized and interaction-sensitive, but several limitations remain. First, our analysis covers only two matched model pairs in the 7B-8B LLM regime. Although the recurrence of the same mid-network region across both pairs is striking, it is not yet clear whether the same pattern holds at other model scales, in denser or sparser architectures, or under different alignment procedures. Extending the transplantation analysis to various model sizes and families would help determine whether the observed localization around layers 8-11 reflects a general property of safety alignment or a scale-specific feature of the models studied in this work.

Moreover, we intentionally constructed filtered subsets to sharpen the contrast between safety-aligned models and base models, but this also limits their sizes. Future work should validate the same localization patterns on larger and less filtered evaluation sets, ideally including benchmarks that jointly measure harmful-prompt refusal and benign-prompt compliance within the same test distribution.

Our greedy forward selection only approximates the space of useful block combinations. The non-monotonic trajectories already show that block effects are not independent, so more exhaustive search strategies may reveal better-performing subsets or more structured interactions. This motivates future studies on combinatorial search, sparse optimization, or learned selection policies that explicitly optimize both malicious refusal and benign non-refusal.

Overall, these limitations and opportunities point to a larger research agenda: using localization not only to diagnose where alignment resides, but also to design safety methods that are more interpretable, more precise, and more robust.

\section{Conclusion}

In this study, we investigate where refusal behavior is encoded in safety-aligned language models by utilizing small aligned–base disagreement prompt sets and transplanting weight matrices into their unaligned counterparts. Our experiments across two model pairs and four benchmarks reveal three consistent empirical patterns that characterize the weight-space localization of safety-aligned refusal behavior:

\begin{itemize}
    \item \textbf{Pathway Concentration:} Transferable refusal behavior is mediated predominantly by the MLP pathway rather than the attention pathway. Replacing MLP weights recovers substantially more refusal behavior, outperforming attention-only hybrids by at least 2.7 times in every malicious benchmark setting we studied.
    \item \textbf{Depth Localization:} Within the MLP stack, a specific mid-depth block (Block 3, layers 8–11) consistently emerges as the strongest localized intervention. It is selected first in all six greedy searches, indicating a disproportionate concentration of refusal-mediating parameters at this depth.
    \item \textbf{Non-Additive Interactions:} Refusal-relevant MLP blocks interact non-monotonically. Adding aligned blocks can reduce malicious-prompt refusal, showing that their behavioral effects are not strictly additive.
\end{itemize}

The findings from MLP-block greedy selection suggest that a selective subset of weights can outperform a full-MLP transplant, achieving a more favorable operating point for both safety and model utility. We also observe a similar pattern from performance of Block~3 layers' transplantation. We emphasize that safety alignment in current LLMs is both \emph{localized} and \emph{interaction-sensitive}. Localization helps explain why aligned behavior can be brittle under subsequent fine-tuning or model modification, while the observed interactions show that effective alignment is not simply a matter of transplanting more aligned parameters. In short, our localization analyses help move from broad behavioral observations toward more targeted safety interventions, directing attention to the components of the network most responsible for refusal behavior.


\section{AI Use Acknowledgment}
We have made use of AI tools (Claude Code, ChatGPT, Google Gemini) to assist with coding and proofreading, editing and formatting during the writing. The human authors take full responsibility for the contents of this paper. 

\section{Funding Acknowledgment}
This work has been supported in part through a grant for the project ``Towards Resilient LLM Alignment: Investigating Brittleness \& Improving Model Safety by Rank Enhancement" funded by the USC Amazon Center on Secure and Trusted Machine Learning

\section{Appendix}
\label{sec:appendix}

Model Generation Configuration for both model families:
\begin{itemize}
    \item Temperature: 0.6
    \item max\_new\_tokens: 1200 (for subset filtering), 2000 (for response generation from hybrid models)
\end{itemize}

\bibliographystyle{plainnat}
\bibliography{references}

\end{document}